# LLMs-Healthcare : Current Applications and Challenges of Large Language Models in various Medical Specialties

Ummara Mumtaz, Awais Ahmed, Summaya Mumtaz

**Abstract**

We aim to present a comprehensive overview of the latest advancements in utilizing Large Language Models (LLMs) within the healthcare sector, emphasizing their transformative impact across various medical domains. LLMs have become pivotal in supporting healthcare, including physicians, healthcare providers, and patients. Our review provides insight into the applications of Large Language Models (LLMs) in healthcare, specifically focusing on diagnostic and treatment-related functionalities. We shed light on how LLMs are applied in cancer care, dermatology, dental care, neurodegenerative disorders, and mental health, highlighting their innovative contributions to medical diagnostics and patient care. Throughout our analysis, we explore the challenges and opportunities associated with integrating LLMs in healthcare, recognizing their potential across various medical specialties despite existing limitations. Additionally, we offer an overview of handling diverse data types within the medical field.

## 1. Introduction

The field of artificial intelligence (AI) has undergone a remarkable evolution in recent years, with significant advancements, particularly noticeable in natural language processing (NLP) and the development of Large Language Models (LLMs). These models represent a paradigm shift in AI's capability to understand, generate, and interact using human language. At their foundation, LLMs are complex algorithms trained on vast, text-based documents and datasets[1]. Such extensive training allows them to recognize patterns adeptly, predict subsequent words in a sentence, and generate coherent, contextually relevant text for the specified inputs, often called prompts within the NLP community. This ability demonstrates the technical prowess of LLMs and signifies their potential to revolutionize how machines understand and process human language. One of the most prominent features of LLMs is their proficiency in processing and analyzing large volumes of text rapidly and accurately, a capability that far surpasses human potential in speed and efficiency[2]. This quality makes them indispensable in areas requiring the analysis of extensive data sets. They are also known as "few-shot" learners, meaning once trained on massive datasets, they can be retrained for new domains utilizing a small number of domain-specific examples[3].

LLMs have become increasingly prevalent in the medical domain, demonstrating their versatility and expanding influence. Their applications in healthcare are multifaceted, ranging from processing vast quantities of medical data and interpreting clinical notes to generating comprehensive, human-readable reports[4]. This broad spectrum of functionalities shows how LLMs are not just tools for data processing but are also instrumental in providing innovative solutions across various aspects of healthcare. LLMs are increasingly being utilized to tackle critical challenges in patient care. This includes providing customized educational content to patients, assisting healthcare professionals in making complex diagnostic decisions, and easing the administrative burdens often associated with healthcare provision[4,5].

While large language models have been applied across a spectrum of activities in healthcare, including medical question answering, examination, pure research-oriented tasks, and administrative duties in hospitals, this review will focus exclusively on their practical applications in healthcare, such as diagnostics and treatment purposes. We uncover their deployment in critical areas such as cancer care, dermatology, dental, and mental health. This exploration is crucial, as it showcases LLMs' capacity to innovate medical diagnostics and patient care, streamline treatment tasks, and address the challenges and opportunities in harnessing their full potential in complex medical areas. We conduct an in-depth analysis of the applications of LLMs across different medical fields, aiming to present a brief yet thorough summary. We focus on the advancements and challenges of integrating these sophisticated models into routine healthcare practices. We offer insights into the current state of progress and identify barriers to their widespread adoption in clinical settings. The paper is structured to cover each medical specialty and associated challenges, followed by examining various data types in the medical field. The conclusion summarizes the findings and implications.

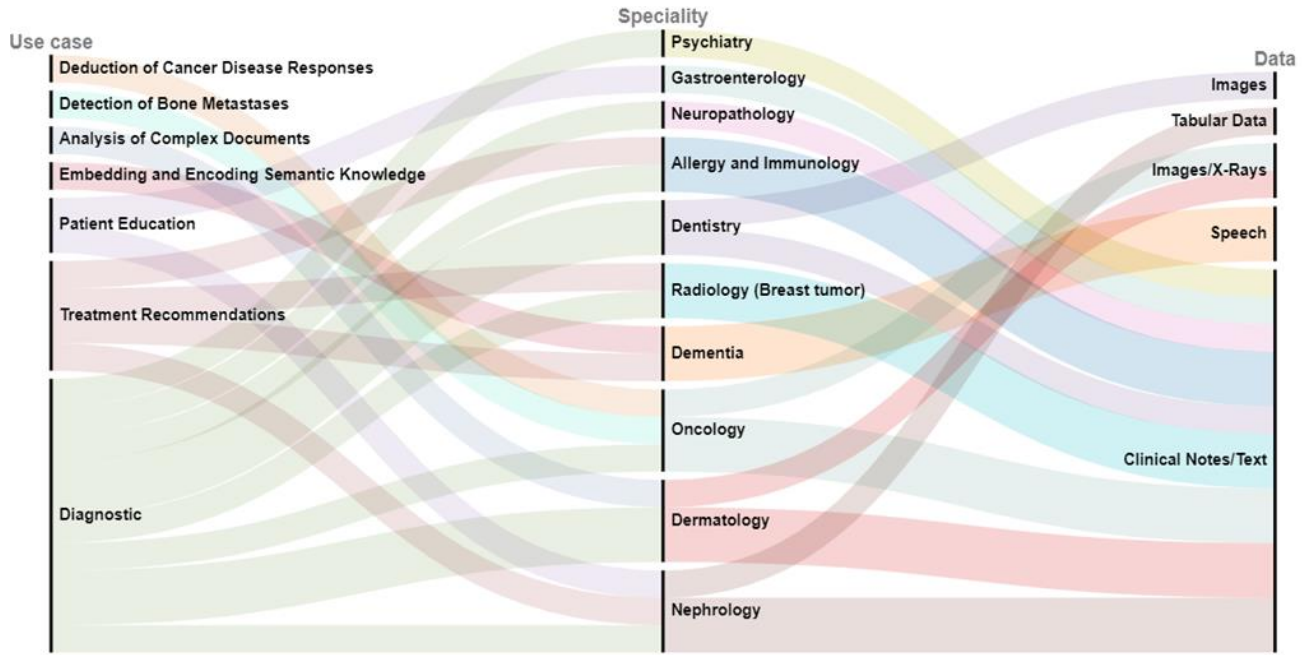

Figure 1 : Visualizing LLM Applications in different medical specialties w.r.t input data type and medical use-case

## 2. Cancer Care (Oncology)

Cancer is characterized by the uncontrolled growth of abnormal cells in the body. It is examined within oncology—studying cancer types and related factors. Adopting Large Language Models (LLMs) such as ChatGPT in oncology has become a focal point of recent research, especially in supporting decision-making processes for cancer treatment. These advanced models are being explored for their capability to enhance diagnostic accuracy, personalize therapy options, and streamline patient care in oncology. By analyzing vast amounts of data, LLMs can provide insights that potentially improve treatment outcomes and patient management strategies. In the subsequent discussion, we explore the studies dedicated to integrating LLMs within oncological care, encapsulating the innovative efforts to harness LLMs' capabilities in enhancing the diagnostic, treatment, and management processes associated with cancer care.

In a study conducted by Vera Sorin and Eyal Klang[6], the capabilities of ChatGPT, a large language model (LLM), were explored as a decision-support tool for breast tumor boards. The research's primary objective was determining how ChatGPT's recommendations align with expert-driven decisions during breast tumor board meetings. For this purpose, clinical data from ten patients discussed in a breast tumor board at their institution was inputted into ChatGPT-3.5. Subsequently, the model's management recommendations were compared with the final decisions made by the tumor board. Moreover, two senior radiologists independently evaluated ChatGPT's responses, grading them on a scale from 1 (complete disagreement) to 5 (complete agreement) across three categories: summarization of the case, the recommendation provided, and the explanation for that recommendation. Most patients in the study, 80%, had invasive ductal carcinoma, with one case each of ductal carcinoma in-situ and a phyllodes tumor with atypia. ChatGPT's recommendations aligned with the tumor board's decisions in seven out of the ten cases, marking a 70% concordance. Upon grading, the first reviewer gave mean scores of 3.7, 4.3, and 4.6 for summarization, recommendation, and explanation, respectively, while the second reviewer's scores were 4.3, 4.0, and 4.3 in the same categories. As an initial exploration, the study suggests that LLMs like ChatGPT could potentially be a valuable asset for breast tumor boards. However, as technology rapidly advances, medical professionals must know its advantages and potential limitations.

In a study by Stefan Lukac and Davut Dayan in January 2023, the capabilities of ChatGPT to assist in the decision-making process for therapy planning in primary breast cancer cases were investigated[7]. Though the ChatGPT was able to identify specific risk factors for hereditary breast cancer and could discern elderly patients requiring chemotherapy assessment for cost/benefit evaluation, it generally offered non-specific recommendations concerning various treatment modalities such as chemotherapy and radiation therapy. Notably, it made errors in patient-specific therapy suggestions, misidentifying patients with Her2 1+ and 2+ (FISH negative) as candidates for trastuzumab therapy and mislabeling endocrine therapy as "hormonal treatment." The study concluded that while ChatGPT demonstrates potential utility in clinical medicine, its current version lacks the precision to offer specific therapy recommendations for primary breast cancer patients. It underscores the necessity for further refinement before it can be a reliable adjunct in multidisciplinary tumor board decisions.

Georges Gebrael assessed the utility of ChatGPT 4.0 to enhance triage efficiency and accuracy in emergency rooms for patients with metastatic prostate cancer[8]. Between May 2022 and April 2023, clinical data of 147 patients presenting with metastatic prostate cancer were examined, of which 56 were selected based on inclusion criteria. ChatGPT demonstrated a high sensitivity of 95.7% for determining patient admissions but had a low specificity of 18.2% for discharges. It agreed with physicians' primary diagnoses in 87.5% of cases. It outperformed physicians regarding accurate terminology usage (42.9% vs. 21.4%) and diagnosis comprehensiveness, having a median diagnosis count of 3 compared to physicians' 2. ChatGPT was more concise in its responses but provided more additional treatment recommendations than physicians. The data suggests that ChatGPT could serve as a valuable tool for assisting medical professionals in emergency room settings, potentially enhancing triage efficiency and the overall quality of patient care.

A study led by Arya Rao et al. investigated the potential of ChatGPT-3.5 and GPT-4 (OpenAI) in aiding radiologic decision-making, specifically focusing on breast cancer screening and breast pain imaging services[9]. The researchers measured the models' responses against the ACR Appropriateness Criteria using two prompt formats: open-ended (OE) and select all that apply (SATA). For breast cancer screening, both versions scored an average of 1.830 (out of 2) in the OE format, but GPT-4 outperformed ChatGPT-3.5 in the SATA format, achieving 98.4% accuracy compared to 88.9%. Regarding breast pain, GPT-4 again showed superiority, registering an average OE score of 1.666 and 77.7% in SATA, while ChatGPT-3.5 scored 1.125 and 58.3%, respectively. The data suggests the growing viability of large language models like ChatGPT in enhancing radiologic decision-making processes, with potential benefits for clinical workflows and more efficient radiological services. However, further refinement and broader use cases are needed for full validation.

Hana et al. conducted a retrospective study in February 2023 to evaluate the appropriateness of ChatGPT's responses to common questions concerning breast cancer prevention and screening[10]. Leveraging methodologies from prior research that assessed ChatGPT's capacity to address cardiovascular disease-related inquiries, the team formulated 25 questions rooted in the BI-RADS Atlas and their clinical experiences within tertiary care breast imaging departments. Each question was posed to ChatGPT three times, and three fellowship-trained breast radiologists critically assessed the responses. The radiologists categorized each response as "appropriate," "inappropriate," or "unreliable" based on the content's clinical relevance and consistency. Their evaluations considered two hypothetical scenarios: content for a hospital website and direct chatbot-patient interactions. The majority's opinion dictated the final determination of appropriateness. Results revealed that ChatGPT provided suitable answers for 88% (22 out of 25) of the questions in both contexts. However, one question pertained to mammography scheduling in light of COVID-19 vaccination, which elicited an inappropriate response.

Additionally, there were inconsistencies in answers related to breast cancer prevention and screening location queries. While ChatGPT frequently referenced guidelines from the American Cancer Society in its responses, it omitted those from the American College of Radiology and the US Preventive Services Task Force. These findings aligned with earlier research by Sarraju et al.[11], where 84% of ChatGPT's cardiovascular disease prevention responses were deemed appropriate. Despite showing considerable potential as an automated tool for patient education on breast cancer, ChatGPT exhibited certain limitations, emphasizing the essential role of physician oversight and the ongoing need for further refinement and research into large language models in healthcare education.

Brian Schulte, in 2023, explored the ability of ChatGPT to identify suitable treatments for advanced solid cancers[12]. Through a structured approach, the study assessed ChatGPT's capacity to list appropriate systemic therapies for newly diagnosed advanced solid malignancies and then compared the treatments ChatGPT suggested with those recommended by the National Comprehensive Cancer Network (NCCN) guidelines. This comparison resulted in the valid therapy quotient (VTQ) measure. The research encompassed 51 diagnoses and found that ChatGPT could identify 91 unique medications related to advanced solid tumors. On average, the VTQ was 0.77, suggesting a reasonably high agreement between ChatGPT's suggestions and the NCCN guidelines. Furthermore, ChatGPT always mentioned at least one systemic therapy aligned with NCCN's suggestions. However, there was a minimal correlation between the frequency of each cancer type and the VTQ. In conclusion, while ChatGPT displays promise in aligning with established oncological guidelines, its current role in assisting medical professionals and patients in making treatment decisions still needs to be defined. As the model evolves, it is hoped that its accuracy in this area will be enhanced, but continued research is essential to fully understand and harness its potential.

In a study led by Julien Haemmerli et al., the capability of ChatGPT was explored in the context of CNS tumor decision-making, specifically for glioma management[13]. Using clinical, surgical, imaging, and immunopathological data from ten randomly chosen glioma patients discussed in a Tumor Board, ChatGPT's recommendations were compared with those of seven CNS tumor experts. While most patients had glioblastomas, findings revealed that ChatGPT's diagnostic accuracy was limited, with a notable discrepancy in glioma classifications. However, it demonstrated competence in recommending adjuvant treatments, aligning closely with expert opinions. Despite its limitations, ChatGPT shows potential as a supplementary tool in oncological decision-making, particularly in settings with constrained expert resources.

In Shan Chen et al.'s research on the effectiveness of ChatGPT in offering cancer treatment advice, the study scrutinized the model's alignment with the National Comprehensive Cancer Network (NCCN) guidelines for breast, prostate, and lung cancer treatments[14]. Through four diverse prompt templates, the study assessed if the mode of questioning influenced the model's responses. While ChatGPT's recommendations aligned with NCCN's guidelines in 98% of the prompts, 34.3% of these recommendations also presented information that needed to be more in sync with the NCCN guidelines. The study concluded that, despite its potential, ChatGPT's performance in consistently delivering reliable cancer treatment advice was unsatisfactory. Consequently, patients and medical professionals must exercise caution when relying on ChatGPT and similar tools for educational purposes.

## 2.1. Challenges associated with LLMs as a decision-support tool in Cancer Care:

While integrating Large Language Models (LLMs) like ChatGPT into oncology shows promise, particularly in decision support for cancer treatment, it also presents several critical challenges, as discussed in the previous section. These challenges must be addressed to ensure LLMs' safe and effective use in high-stakes medical environments. Firstly, the issue of accuracy and precision in LLMs is a significant concern. For instance, in Julien Haemmerli's[13] study on glioma therapy, ChatGPT demonstrated limitations in accurately classifying glioma types. Similarly, the study by Stefan Lukac and Davut Dayan[7] revealed errors in patient-specific therapy suggestions, such as misidentifying patients for trastuzumab therapy. These inaccuracies highlight the risk of potential misdiagnoses or inappropriate treatment recommendations, which could have profound implications for patient care.

Another challenge is the capacity of LLMs to consider the comprehensive clinical picture, including patient functional status, which is often a nuanced judgment call made by experienced physicians. ChatGPT's moderate performance in this area, as seen in Haemmerli's study[13], indicates a gap between current LLM capabilities and the complex decision-making processes in medical practice. Furthermore, the integration of LLMs into existing medical workflows raises concerns. For example, Georges Gebrael's[8] study on triage in metastatic prostate cancer showed that while ChatGPT had high sensitivity, its low specificity for discharges could lead to operational inefficiencies. Integrating LLMs within healthcare systems also poses challenges in data privacy, interoperability, and the need for robust IT infrastructure.

Lastly, the role of LLMs in patient education and communication is not without limitations. Hana L Haver et al.[10] studies demonstrated inconsistencies in ChatGPT's responses to breast cancer prevention and screening questions. This inconsistency highlights the importance of human oversight in verifying the information provided by LLMs, ensuring it aligns with established medical guidelines and practices. In summary, while LLMs present exciting opportunities for enhancing cancer care, their current limitations in accuracy, comprehensive clinical assessment, integration into existing systems, and patient education necessitate a cautious and critical approach. These models should be viewed as supplementary tools that augment, rather than replace, the expertise of medical professionals. Continuous evaluation, refinement, and ethical consideration are essential to harness the full potential of LLMs in oncology.

## 3. Skin Care: Dermatology

Our skin is a barrier against external threats such as viruses, bacteria, and other harmful organisms. Dermatology is the branch of medicine dealing with skin diseases. There has been a surge in cases related to skin diseases in the past years, affecting people of all ages[15]. Common skin-related diseases include acne, alopecia, bacterial skin infections, decubitus ulcers, fungal skin diseases, pruritus, and psoriasis[16]. Traditional dermatology diagnosis is based on a visual inspection of skin features and subjective evaluation by a dermatologist[17]. The realm of dermatology diagnosis faces several significant challenges. Firstly, accurately interpreting skin disease imagery is complex due to the wide variety of skin conditions and their subtle visual differences. This task requires a high level of expertise, leading to the second challenge: a noticeable shortage of dermatologists, especially in remote or underserved areas. Lastly, creating patient-friendly diagnostic reports is another hurdle. These reports need to be detailed yet understandable to non-specialists, making their production time-consuming and labor-intensive for dermatologists.

In addressing the above challenges in dermatological diagnostics, Zhou et al. introduced SkinGPT-4, an innovative interactive dermatology diagnostic system underpinned by an advanced visual Large Language Model[18]. This study was mainly focused on tackling the prevalent issues in dermatology, such as the shortage of specialized medical professionals in remote areas, the intricacies involved in interpreting skin disease images accurately, and the demanding nature of creating patient-friendly diagnostic reports. SkinGPT-4, utilizing a refined version of MiniGPT-4, trained on an extensive dataset that included 52,929 images of skin diseases, both from public domains and proprietary sources, along with detailed clinical concepts and doctors' notes. This comprehensive training on skin-related disease images enabled SkinGPT-4 to articulate medical features in skin disease images using natural language and make precise diagnoses. The functionality of SkinGPT-4 allows users to upload images of their skin conditions, after which the system autonomously analyzes these images. It identifies the characteristics and categorizes the skin conditions, performs an in-depth analysis, and provides interactive treatment recommendations. A notable aspect of SkinGPT-4 is its local deployment feature, combined with a solid commitment to maintaining user privacy, making it a viable option for patients seeking accurate dermatological assessments. To ascertain the efficacy of SkinGPT-4, the study conducted a series of quantitative evaluations on 150 real-life dermatological cases. Certified dermatologists independently reviewed these cases to validate the diagnoses provided by SkinGPT-4. Among the 150 cases, a commendable 78.76% of the diagnoses rendered by SkinGPT-4 were validated as either accurate or relevant by the dermatologists, breaking down into 73.13% that firmly aligned and another 5.63% that agreed. The outcomes of this evaluation underscored the accuracy of SkinGPT-4 in diagnosing skin diseases. While SkinGPT-4 is not positioned as a replacement for professional medical consultation, its contribution to enhancing patient comprehension of medical conditions, improving communication between patients and doctors, expediting dermatologists' diagnostic processes, and potentially fostering human-centered care and healthcare equity in underdeveloped regions is significant.

## 3.2. Challenges associated with utilizing LLMs in Dermatology:

The introduction of SkinGPT-4 by Zhou et al. marks a significant advancement in dermatological diagnostics, addressing challenges like the dermatologist shortage and the complexities of skin disease image interpretation and patient-friendly report generation[18]. Despite its innovative approach and the training on an extensive dataset to articulate medical features in skin images, there are inherent challenges. Some challenges associated with deploying SkinGPT-4 include ensuring consistent diagnostic accuracy across various skin conditions, safeguarding patient privacy while managing sensitive health data, and integrating the technology seamlessly into existing healthcare systems. Additionally, despite SkinGPT-4's high diagnostic accuracy, continuous human oversight in medical diagnosis and treatment planning remains critical to complement the AI's capabilities with professional medical judgment and ensure optimal patient care outcomes. Additionally, advancements might focus on developing models that can adapt to new, emerging skin conditions and leveraging telemedicine to extend dermatological care to remote areas, thus promoting healthcare equity.

## 4. Neurodegenerative Disorders: Dementia & Alzheimer's

Neurodegenerative disorders involve the gradual deterioration of specific neuron groups, differing from the non-progressive neuron loss seen in metabolic or toxic conditions. These diseases are categorized by their primary symptoms (such as dementia, parkinsonism, or motor neuron disease), the location of neurodegeneration within the brain (including frontotemporal degenerations, extrapyramidal disorders, or spinocerebellar degenerations), or by the underlying molecular abnormalities[19]. Dementia is a broad category of brain diseases that cause a long-term and often gradual decrease in the ability to think and remember, affecting daily functioning. Alzheimer's disease (AD) is the most common cause of dementia, characterized by memory loss, language problems, and unpredictable behavior.

LLMs such as Google Bard and ChatGPT have emerged as valuable tools for predicting neurodegenerative disorders. A study by Koga et al. evaluated these models' predictive accuracy using cases from Mayo Clinic conferences[20]. Using the Mayo Clinic brain clinicopathological conferences as their sample pool, the researchers extracted 25 cases of neurodegenerative disorders. These clinical summaries were then utilized for training and testing the models. The diagnoses offered by each model were compared against the official diagnosis provided by medical professionals. Findings from the study highlighted that ChatGPT-3.5 aligned with 32% of all the physician-made diagnoses, Google Bard with 40%, and ChatGPT-4 with 52%. When assessing the accuracy of these diagnostic predictions, ChatGPT-3.5 and Google Bard both achieved a commendable score of 76%, while ChatGPT-4 led the pack with an impressive accuracy rate of 84%. The evident proficiency exhibited by LLMs, specifically ChatGPT and Google Bard, highlights their considerable potential in revolutionizing diagnostic processes in neurodegenerative disorders.

This study conducted by Agbavor and Liagn (2022) explored the use of GPT-3-generated text embeddings to predict dementia, utilizing data from the ADReSSo Challenge (Alzheimer's Dementia Recognition through Spontaneous Speech *only* challenge[22]), which focuses on identifying cognitive impairment through spontaneous speech[21]. The author proposed using the model to identify individuals with dementia against healthy individuals as controls. Using the 237 speech recordings derived from the ADReSSO (Alzheimer's Dementia Recognition through Spontaneous Speech *only* challenge), the author used a 70/30 split and obtained 71 data samples as the testing set and 166 as the training set. In the training set, 87 individuals had AD, and 79 were healthy controls. GPT-3 was innovatively used for embedding the transcribed speech texts. Then, the model extracts the acoustic features such as temporal analysis (periodicity of speech, pause rate, phonation rate, etc.) and speech production (vocal quality, articulation, prosody, etc.). These features serve as the input for the classification model used in AD prediction. GPT-3 embeddings are then compared with BERT and traditional acoustic features. The findings reveal that text embeddings outperform traditional acoustic methods and compare well with fine-tuned models such as BERT. This suggests that GPT-3's text embeddings offer a promising approach for early dementia diagnosis.

Another study conducted by Mao and colleagues[23] outlines developing and applying a deep learning framework utilizing the BERT model for predicting the progression from Mild Cognitive Impairment (MCI) to Alzheimer's Disease (AD) using unstructured EHR notes. The study cataloged 3,657 MCI-diagnosed patients and their clinical notes from Northwestern Medicine Enterprise Data Warehouse (NMEDW) between 2000 and 2020, using only their initial MCI diagnosis notes for analysis. These notes underwent de-identification, cleaning, and segmentation before training an AD-specific BERT model (AD-BERT). AD-BERT transformed patient note sections into vector forms, which a fully connected network analyzed to predict MCI-to-AD progression. For validation, a similar methodology was applied to 2,563 MCI patients from Weill

Cornell Medicine (WCM). AD-BERT outperformed seven baseline models, showing superior accuracy in both patient groups, evidenced by its AUC and F1 scores.

In the diagnosis of complex conditions like Alzheimer's disease, medical professionals use a variety of data such as images, patient demographics, genetic profiles, medication history, cognitive assessments, and speech data. Some of the recent studies have proposed multi-modal AD diagnosis or prediction methods leveraging the popular pre-trained large language model (LLM) to add text data sources, in addition to images and other data types[24,25-26].

### 4.1. Challenges associated with LLMs in Neurodegenerative disorders

Utilizing LLMs in diagnosing and managing neurodegenerative disorders like dementia and Alzheimer's disease presents several challenges. Firstly, the complexity and variability of these conditions require highly accurate and deep understanding, which LLMs may not always provide due to limitations in their training data. The ethical and privacy concerns about handling sensitive patient data pose significant hurdles. Furthermore, integrating these models into clinical workflows demands substantial validation to ensure they complement, rather than complicate, healthcare professionals' decision-making processes. Lastly, there's a need for continuous updates and improvements in these models to keep pace with the latest medical research and clinical practices

## 5. Dentistry

The World Health Organization reports that oral diseases impact approximately 3.5 billion individuals globally, with dental caries, periodontal diseases, and tooth loss being the most prevalent. These conditions, largely preventable and manageable with early diagnosis, have seen the application of AI methodologies in recent years, including the diagnosis of dental caries[27, 28] and periodontitis[29]. Despite this, exploring Large Language Models (LLMs) in dentistry remains notably scarce, with limited studies demonstrating their practical application.

Huang et al. stand out by proposing LLM-based deployment strategies within dentistry, marking the emerging area of research with significant potential for advancement[29]. To showcase the effectiveness and potential of applying Large Language Models (LLMs) in dentistry, this work introduced a framework for an automated diagnostic system utilizing Multi-Modal LLMs. This innovative system incorporated three distinct input modules: visual, auditory, and textual data, enabling comprehensive analysis. Visual inputs, such as dental X-rays and CT scans, are evaluated for anomalies using vision-language models, facilitating precise diagnostics. Audio inputs serve dual purposes: detecting voice anomalies and understanding patient narratives, which are converted to text for further analysis by LLM. To illustrate the capabilities of the multi-modal LLM AI system in dental practice, Huang et al. proposed its application in diagnosing and planning treatment for dental caries. The process begins with inputting a tooth's X-ray into the system, where vision-language modeling is employed to detect any decay on the tooth. Once identified, the system utilizes LLM to propose a comprehensive treatment plan, articulated through seven detailed steps. These steps range from initial patient communication to scheduling follow-up appointments, highlighting a thorough approach to patient care. Despite its advanced diagnostics, the system's limitations, such as failing to detect potential bone loss, are acknowledged, suggesting areas for further research and development to enhance its effectiveness in dental diagnostics.

**5.1. Challenges associated with dental care:**

The accuracy of LLMs like ChatGPT depends on the availability of high-quality, relevant dental data. A significant hurdle in designing and training LLMs for dental care is limited access to the dental records owned by private dental clinics and concerns over patient privacy, which restricts access to comprehensive and current datasets. LLMs' development and effectiveness in dentistry must navigate these challenges, ensuring access to extensive, up-to-date information while addressing privacy and ownership issues to avoid biases and maintain data integrity.

The potential of LLMs in dental healthcare seems promising and can revolutionize how dental professionals diagnose, treat, and manage patient care today. LLMs could significantly improve diagnostic precision by leveraging the vast amounts of data available in patient records and imaging, allowing for early detection and intervention in dental conditions. Furthermore, the ability of LLMs to generate personalized treatment plans and educational materials tailored to individual patient needs could enhance the effectiveness of patient care. This personalization and the model's ability to process and analyze data swiftly could lead to more efficient and patient-centered dental healthcare practices. As LLMs continue to evolve, their integration into dental healthcare is expected to deepen, offering innovative solutions to longstanding challenges and improving patient outcomes worldwide.

**6. Mental Health: Psychiatry and Psychology**

Mental health disorders, which affect millions globally, significantly reduce the life quality of individuals and their families. In psychiatry, LLMs have the potential to refine diagnostic precision, optimize treatment outcomes, and enable more tailored patient care, moving beyond traditional, subjective diagnostic approaches prone to inaccuracies. By leveraging AI to analyze extensive patient data, it's possible to uncover patterns not easily detectable by humans, thereby improving diagnosis[28,29].

Galatzer-Levy and colleagues, in 2023, delved into exploring the potential role of large language models (LLM) in psychiatry[30]. Their primary investigation tool was Med-PALM 2, an LLM equipped with comprehensive medical knowledge. The model was trained and tested using a blend of clinical narratives and patient interview transcripts. The dataset encompassed expert evaluations using instruments like the 8-item Patient Health Questionnaire and the PTSD Checklist-Civilian Version (PCL-C). The study intended to gauge the severity of PTSD using the PCL-C while employing the PHQ-8 to assess depression and anxiety levels. The evaluation process involved extracting from Med-PALM 2 clinical scores, the rationale for such scores, and the model's confidence in its derived results. The gold standard for this evaluation was the DSM 5 (Diagnostic and Statistical Manual of Mental Disorders, Fifth Edition). The researchers' rigorous testing process involved the analysis of 46 clinical case studies, 115 PTSD evaluations, and 145 depression instances. These were probed using prompts to tease out diagnostic information and clinical scores. The rigorous assessment also saw Med-PaLM 2 fine-tuned through many natural language applications and a substantial textual database. Notably, research-quality clinical interview transcripts were employed as inputs when assessing the model's efficacy. Med-PaLM 2 demonstrated its prowess in evaluating psychiatric states across various psychiatric conditions. Remarkably, when tasked with predicting psychiatric risk from clinician and patient narratives, the model showcased an impressive accuracy rate ranging between 80% and 84%.

Another study evaluated the performance of various LLMs, including Alpaca and its variants, FLAN-T5, GPT-3.5, and GPT-4, across different mental health prediction tasks such as mental state (depressed, stressed or risk actions like suicide) using online text[31]. Through extensive experimentation, including zero-shot, few-shot, and instruction fine-tuning methods, it was found that instruction fine-tuning notably enhances LLMs' effectiveness across all tasks. Notably, the fine-tuned models, Mental-Alpaca and Mental-FLAN-T5, demonstrated superior performance over larger models like GPT-3.5 and GPT-4 and matched the accuracy of task-specific models.

The use of conversational agents based on LLMs for mental well-being support is growing, yet the effects of such applications still need to be fully understood. A qualitative study by Ma et al. of 120 Reddit posts and 2,917 comments from a subreddit dedicated to mental health support apps like Replika reveals mixed outcomes[32]. While Replika offers accessible, unbiased support that can enhance confidence and self-exploration, it struggles with content moderation, consistent interactions, memory retention, and preventing user dependency, potentially increasing social isolation.

Following the advancements with ChatGPT, research into automated therapy using AI's latest technologies is gaining momentum. This new direction aims to shift mental health assessments from traditional rating scales to a more natural, language-based communication. The emergence of large language models, like those powering ChatGPT and BERT, marks a significant shift in artificial intelligence, potentially revolutionizing standardized psychological assessments. This evidence points towards AI's capacity to transform mental health evaluations into interactions that mirror natural human communication, pending comprehensive validation in specific application scenarios[33].

### 6.1. Challenges associated with applications of LLMs for Mental Health

In mental health applications, LLMs face challenges like ensuring content sensitivity and safety to avoid harmful advice, maintaining accuracy and reliability to prevent misdiagnoses, and offering personalized, empathetic responses for adequate support. Data privacy and security are paramount due to the personal nature of discussions. There's also a need to prevent user over-reliance on LLMs, potentially deterring professional help. Ethical considerations include the impact of replacing human interactions with AI and avoiding biases. Additionally, navigating regulatory compliance within mental health laws and guidelines is crucial for lawful operation.

### 8   Other Medical Specialties: Nephrology, Gastroenterology, Allergy and immunology

The integration of Large Language Models into medical specialties like nephrology and gastroenterology remains in the early stages, with their full potential yet to be realized. Current applications in these areas are sparse, highlighting opportunities for future exploration and implementation. This brief overview aims to shed light on the existing implementations of LLMs within these specific fields, indicating the nascent but promising role of advanced AI technologies in enhancing diagnostic and treatment methodologies in nephrology and gastroenterology.

## 8.1. Nephrology

Within the domain of nephrology, LLMs are being utilized to assist in diagnosing kidney diseases, providing treatment guidance, and monitoring renal function, as noted by Wu and colleagues[34]. These LLMs facilitate the evaluation of crucial data such as laboratory results, clinical data, and a patient's medical history during the diagnostic phase. As such, the LLMs chosen for nephrological applications are often preferred to possess a sophisticated medical knowledge capability, especially in multiple-choice medicine test-taking. Various LLMs, including Orca Mini 13B, Stable Vicuna 13B, Falcon 7B, Koala 7B, Claude 2, and GPT-4, have found applications in treating and diagnosing kidney diseases. However, owing to their unique zero-shot reasoning capabilities, GPT-4 and Claude 2 are particularly suitable for this intricate medical specialty. Currently, these models are employed to respond to multiple-choice questions about nephrology. Wu et al. incorporated questions from clinical backgrounds linked to 858 nephSAP multiple-choice queries collated between 2016 and 2023. When evaluating the proficiency of Claude 2 and GPT-4, performance was gauged based on the proportion of correctly answered nephrology-related nephSAP multiple-choice questions. GPT-4 demonstrated superior performance, garnering a score of 73.3%, in contrast to Claude 2, which achieved a score of 54.4%. When individual nephrology topics were examined, GPT-4 consistently outperformed its counterparts, including Claude 2, Vuna, Kaola, Orca-mini, and Falcon.

## 8.2. Gastroenterology

Lahat et al. explored the capabilities of large language models, specifically OpenAI's ChatGPT, in responding to queries within the realm of gastrointestinal health[35]. Their evaluation employed 110 real-world questions, benchmarking ChatGPT's responses against the expert consensus of seasoned gastroenterologists. These queries spanned a spectrum of topics, from diagnostic tests and prevalent symptoms to treatments for a range of gastrointestinal issues. The source of these questions was public internet platforms. The researchers evaluated the outputs of ChatGPT on metrics such as accuracy, clarity, up-to-dateness, and efficacy, rating them on a scale from 1 to 5. These outputs were then categorized into symptoms, diagnostic tests, and treatments. ChatGPT averaged scores of 3.7 for clarity, 3.4 for accuracy, and 3.2 for efficacy in the symptom category. Diagnostic test-related queries resulted in scores of 3.7 for clarity, 3.7 for accuracy, and 3.5 for efficacy. As for treatment-related questions, the model achieved 3.9 for clarity, 3.9 for accuracy, and 3.3 for efficacy. The results indicated the substantial potential of ChatGPT in providing valuable insights within the gastrointestinal specialty.

### 8.3. Allergy and immunology

In allergy and immunology, LLMs akin to their applications in dermatology, have shown promising potential. According to a study by Goktas et al., LLMs, specifically models like GPT-4 and Google Med-PaLM2, significantly enhance the diagnostic process within allergy and immunology disciplines[36]. These advanced models elevate the precision of diagnosis and can tailor treatment plans to suit individual patient needs. Beyond the clinical realm, they also play a pivotal role in fostering patient engagement, ensuring patients are actively involved and informed in their healthcare journey. As a result, the integration of LLMs in allergy and immunology represents a paradigm shift towards more accurate, personalized, and patient-centric medical care.

**Section 9: Handling different types of data in the medical industry**

This section provides an overview of how different data formats and types are handled in the medical industry when used as training data or inputs for a large language model.

### 9.1. Clinical Notes

Clinical notes, an integral component of patient health records, have increasingly been utilized in medicine as input to large language models (LLMs). These notes, typically generated by healthcare professionals, serve as rich patient information repositories, including their medical history, present symptoms, diagnoses, treatments, and more. Clinical notes are fed into LLMs to extract meaningful patterns, predictions, and insights. Before using these notes, they are often preprocessed to ensure they are in a format that's easily digestible for the models. This preprocessing can involve converting handwritten notes into digital formats, anonymizing patient data to maintain privacy, and structuring the data in a consistent format. LLMs can directly process these notes and produce a range of tools suited for activities like condensing medical data, assisting in clinical decisions, and creating medical reports[37]. To utilize clinical notes in LLMs, prompts containing questions, scenarios, or comments about the note are used, such as "Assume the role of a neurologist at the Mayo Clinic brain bank clinicopathological conference." Based on this, the model provides an output that aids in evaluation or diagnosis across different medical fields[37].

## 9.2. X-rays/ Images

X-rays are medical imaging that utilizes ionizing radiation to produce images of internal body organs. This data type may include CT scans (tomography), chest X-rays, and bone X-rays. In medicine, X-ray images can be processed by a computer-aided detection (CAD) model, which is pre-trained to derive the outputs in tensor form. These tensors are then translated into natural language, where they can be used as LLM input to generate summaries or descriptions of the X-ray images. Wang et al. illustrated how the X-rays of exam images are handled for utilizing them with the LLMs[38]. They established that the model is fed into pre-trained CAD models to derive the output. Then, translate the tensor (output) into natural language. Lastly, the language models are used to make conclusions and summarize the results. They establish that X-ray images can be used as input in the LLM and fed into the model with prompts to generate the image summarization or descriptive caption. The LLM supports visual question answering, where the x-ray images of the patients are fed into an image encoder (BLIP-2), where the natural language presentation is generated and embedded based on the image understanding.

Bazi and colleagues proposed a transformer encoder-decoder architecture to handle the visual data when using the LLM[39]. They extracted the image features using the vision transformer (ViT) model and then used the textual encoder transformer to embed the questions. It is then fed to the resulting textual and visual representations into a multi-modal decoder to generate the answers. To demonstrate how LLM handles the visual data, they used VQA datasets for radiology images, termed PathVQA and VQA-RAD. In decoding the radiology images, the proposed model achieved 72.97% and 8.99% for the VQA-RAD and 62.37% or 83.86% for PathVQA.

## 9.3. Radiological reports

Radiological reports are documents from radiologists that present the findings or interpretation of medical imaging studies such as MRIs, X-rays, and CT scans. These data are processed as texts within the report to be input for LLMs in medicine. After data augmentation, the radiological reports are used as inputs in the LLM model. Tan and colleagues collected 10,602 CT scan reports from patients with cancer at a single facility[40]. These reports were categorized into four response types: no evidence of disease, partial response, stable disease, or progressive disease. To analyze these reports, we utilized various models, including transformer models, a bidirectional LSTM model, a CNN model, and traditional machine learning approaches. Techniques such as data augmentation through sentence shuffling with consistency loss and prompt-based fine-tuning were applied to enhance the performance of the most effective models.

## 9.4. Speech data

Speech data, encompassing medical interviews, consultations, and patient audio interactions, serves as a valuable reservoir of information. Before its use in Large Language Models (LLMs), this data is converted into a textual format through automatic speech recognition (ASR) systems. Notably, converting audio data into text is accomplished using pre-trained models, with Wav2vec 2.0 emerging as a leading contender in speech recognition technology. In their groundbreaking work, Agbavor and Liang[21] employed the Wav2vec2-base-960 base model, an advanced tool fine-tuned on an extensive 960-hour dataset of 16 kHz speech audio. Their methodology incorporated Librosa for audio file loading and Wav2Vec2Tokenizer for the crucial task of waveform audio tokenization. These tokenized audio segments are inputted into the Wav2Vec2ForCTC model depending on memory capacities. This model decodes the tokens, resulting in the generation of text transcripts. Furthermore, an alternative approach to leveraging speech data in LLMs involves using open MILE, an open-source toolkit. Open MILE offers functionalities like speech classification and facilitates extracting audio features from speech or musical signals, proving its versatility in handling audio data for various applications.

## 9.5. Tabular Data

In the medical domain, tabular data typically encompasses clinical measurements, patient records, and lab outcomes, arranged methodically in a matrix of rows and columns. A transformation via tabular modeling is requisite for this structured data to be effectively utilized by Large Language Models (LLMs). The ubiquity of this tabular format in clinical and physician databases has often led to the use of tree-based models like bagging and boosting. However, these models come with their share of limitations. Highlighting an innovative approach to this challenge, Chen et al. presented a study employing a data set of 1479 patients undergoing immune checkpoint blockade (ICB) treatments for various cancer types[41]. Segmenting the dataset, with 295 patients for testing and 1184 for training, they unveiled how LLMs process tabular data. Crucial to this process is serializing the feature columns into coherent sequences of natural language tokens that the LLM can interpret. This serialization can be achieved through various methods, be it the prompting-based regeneration approach, using {attribute} is {value} functions, or manual serialization templates.

Furthermore, Chen and his team introduced an advanced tabular model, ClinTaT, augmented from its original design. This refined model incorporates a continuous embedding layer harmonized with multiple distinct layers that mirror the table's continuous feature count. Continuous variables are melded with embedded categorical data for the final processing step, which is then channeled into the transformer for analysis.

## 10. Conclusion

Large Language Models (LLMs) applications have carved out a transformative niche in the healthcare sector. From patient engagement and education to diagnostic assistance, administrative support, and medical research, the multifaceted applications of LLMs have demonstrated their potential to optimize various facets of the medical landscape. Their expansive knowledge repositories and adeptness at understanding context and generating human-like textual responses have positioned LLMs as invaluable assets within the healthcare domain. Their integration with chatbots offers a more personalized and efficient patient experience, aiding in tasks ranging from medication clarification to mental health support. On the diagnostic front, incorporating LLMs with electronic health systems and medical imaging promises to enhance the accuracy and efficiency of diagnosis and treatment plans. LLMs' capability to assist in clinical documentation, medical language translation, and medical education for patients highlights their adaptability and relevance in varied healthcare scenarios.

However, while the benefits of LLMs are numerous, their practical application in the healthcare sector also underscores the importance of precision, context awareness, and ethical considerations, given the critical nature of medical decision-making. While LLMs like ChatGPT and Med-PaLM have shown significant potential, there's an imperative for ongoing refinement, especially when handling complex or rare medical cases. As LLMs become more integrated into patient care, research addressing the ethical implications, including data privacy, the balance between automation and human intervention, and informed patient consent, will be paramount. Collaborative research exploring the fusion of LLMs with other emerging technologies, such as augmented reality or wearable health devices, can open new avenues for patient care and remote monitoring. Enhancing the LLMs' contextual understanding is crucial. Future work should focus on the model's ability to consider a patient's medical history and present conditions before offering recommendations. In sum, the horizon of LLMs in healthcare is expansive and promising. As we continue to witness the convergence of technology and medicine, the collaboration of multidisciplinary teams—combining expertise from AI, medicine, ethics, and other domains—will be integral to harnessing the full potential of LLMs in healthcare.

31. Xu X, Yao B, Dong Y, et al., 2023, Leveraging large language models for mental health prediction via online text data. arXiv preprint arXiv:2307.14385.
32. Ma Z, Mei Y, Su Z., 2024, Understanding the Benefits and Challenges of Using Large Language Model-based Conversational Agents for Mental Well-being Support. *AMIA Annu Symp Proc.* 11;2023:1105-1114.
33. Kjell, O., Kjell, K., & Schwartz, H. A., 2023, AI-based large language models are ready to transform psychological health assessment.
34. Wu S, Koo M, Blum, et al., 2023, A comparative study of open-source large language models, GPT-4 and Claude 2: Multiple-choice test taking in nephrology. arXiv.org. https://arxiv.org/abs/2308.04709
35. Lahat A, Shachar E, Avidan B, et al., 2023, Evaluating the utility of a large language model in answering common patients' gastrointestinal health-related questions: Are we there yet? *Diagnostics*, 13:1950. https://doi.org/10.3390/diagnostics13111950
36. Goktas P, Karakaya G, Kalyoncu, et al., 2023, Artificial intelligence chatbots in allergy and immunology practice: Where have we been and where are we going? The *Journal of Allergy and Clinical Immunology: In Practice*, 11:2697-2700. https://doi.org/10.1016/j.jaip.2023.05.042
37. Singhal K, Azizi S, Tu T, et al., 2023, Large language models encode clinical knowledge. https://www.nature.com/articles/s41586-023-06291-2.
38. Wang S, Zhao Z., Ouyang, X., et al., 2023, ChatCAD: Interactive Computer-Aided Diagnosis on Medical Image using Large Language Models. *Computer Science*, 1-11. https://arxiv.org/abs/2302.07257
39. Bazi Y, Rahhal MM, Bashmal L, Zuair M, 2023, Vision–language model for visual question answering in medical imagery. *Bioengineering*, 10(3), 380. https://doi.org/10.3390/bioengineering10030380
40. Tan RS, Lin Q, Low GH, et al., 2023, Inferring cancer disease response from radiology reports using large language models with data augmentation and prompting. *Journal of the American Medical Informatics Association*, 1-8. https://doi.org/10.1093/jamia/ocad133.
41. Chen, Z., Balan, M. M., & Brown, K., 2023, Language models are few-shot learners for prognostic prediction. arXiv.org. https://arxiv.org/abs/2302.12692
24